\documentclass[11pt,a4paper]{article}
\usepackage{authblk}
\usepackage[hyperref]{acl2020}

\usepackage{url}
\usepackage{graphicx}
\usepackage{amsmath}
\usepackage{amssymb}
\usepackage{pgfplots}
\pgfplotsset{compat=1.8}
\usepackage{mathtools}
\usepackage{tikz}
\usepackage{wrapfig}
\usepackage{caption}
\usepackage{subcaption}
\usepackage{cleveref}
\usepackage{tabularx}

\usepackage{algorithm}
\usepackage[noend]{algpseudocode}
\usepackage{relsize}
\usepackage{multirow}
\usetikzlibrary{fit,calc}
\usepackage{xcolor} 
\usepackage{pifont}
\usepackage{arydshln}

\usepackage{booktabs}

\usepackage{microtype}

\aclfinalcopy % Uncomment this line for the final submission

\graphicspath{{./figures/}}

\title{A Hierarchical Decoder with Three-level Hierarchical Attention to Generate Abstractive Summaries of Interleaved Texts}

\author[1,3]{\bf Sanjeev Kumar Karn}
\author[2]{\bf Francine Chen}
\author[2]{\bf Yan-Ying Chen}
\author[3]{\bf Ulli Waltinger}
\author[1]{\bf Hinrich Sch\"{u}tze}
\affil[1]{Center for Information and Language Processing (CIS), LMU Munich}
\affil[2]{FX Palo Alto Laboratory, Palo Alto, California}
\affil[3]{Machine Intelligence, Siemens CT, Munich, Germany}
\affil[1]{\tt skarn@cis.lmu.de}
\affil[2]{\tt \{chen,yanying\}@fxpal.com}

\date{}
\def\figlabel#1{\label{fig:#1}\label{p:#1}}
\def\figref#1{Figure~\ref{fig:#1}}
\def\eqref#1{Eq.~\ref{eqn:#1}}

\def\eqlabel#1{\label{eqn:#1}}

\def\tabref#1{Table~\ref{tab:#1}}
\def\tablabel#1{\label{tab:#1}\label{p:#1}}

\newcommand\boldblue[1]{\textcolor{blue}{\textbf{#1}}}
\newcommand\boldgreen[1]{\textcolor{green}{\textbf{#1}}}
\newcommand\boldred[1]{\textcolor{red}{\textbf{#1}}}
\newcommand\boldyellow[1]{\textcolor{yellow}{\textbf{#1}}}
\newcommand{\tikzmark}[1]{\tikz[overlay,remember picture] \node (#1) {};}
\newcommand{\DrawBox}[3][]{%
    \tikz[overlay,remember picture]{
    \draw[black,#1]
      ($(#2)+(-0.5em,2.0ex)$) rectangle
      ($(#3)+(0.75em,-0.75ex)$);}
}
\newcommand\Tstrut{\rule{0pt}{2.6ex}}  

\def\algref#1{Algorithm.~\ref{alg:#1}}
\algdef{SE}[VARIABLES]{Variables}{EndVariables}
   {\algorithmicvariables}
   {\algorithmicend\ \algorithmicvariables}
\algnewcommand{\algorithmicvariables}{\textbf{global}}

\begin{document}
%\author{Anonymous Submission}
\maketitle
\begin{abstract}
Interleaved texts, where posts belonging to different threads occur in one sequence, are a common occurrence, e.g., online chat conversations. To quickly obtain an overview of such texts, existing systems first disentangle the posts by threads and then extract summaries from those threads. The major issues with such systems are error propagation and non-fluent summary. To address those, we propose an end-to-end trainable hierarchical encoder-decoder system. We also introduce a novel hierarchical attention mechanism which combines three levels of information from an interleaved text, i.e, posts, phrases and words, and implicitly disentangles the threads. We evaluated the proposed system on multiple interleaved text datasets, and it out-performs a SOTA two-step system by 20-40\%. 
\end{abstract}

\section{Introduction}
\begin{figure}[t!]
\footnotesize
\begin{center}
\resizebox{0.85\linewidth}{!}{
\includegraphics[width=\textwidth]{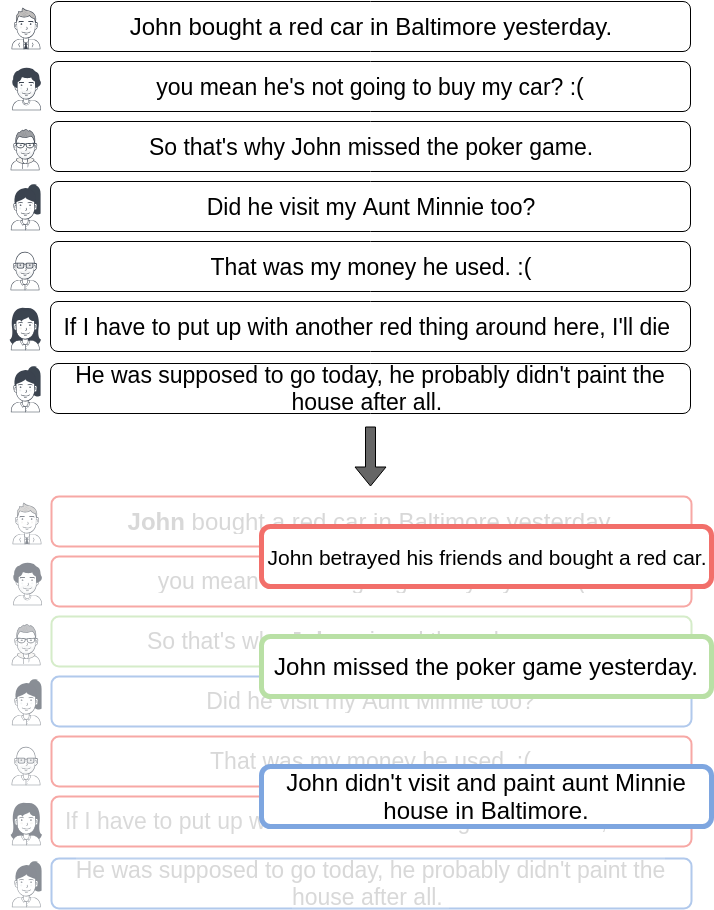}
}
\end{center}
\caption{\label{fig:interlv_summ}In the upper part, 7 interleaved posts belonging to different threads occur in a sequence. In the background at the bottom, posts are disentangled (clustered) into 3 threads (posts are outlined with colors corresponding to threads), and in the foreground, single sentence abstractive summaries are generated for each thread. %Our proposed system takes interleaved posts as input, disentangles the threads implicitly and generates the summaries.
}
\end{figure}

Interleaved texts, e.g., multi-author entries for activity reports, and social media conversations, such as Slack are increasingly common. %, thanks to the new forms of communications.
However, getting a quick sense of different threads in interleaved texts is often difficult due to entanglement of threads, i.e, posts belonging to different threads occurring in one sequence; see a hypothetical example in \figref{interlv_summ}.

In conversation disentanglement, interleaved posts are grouped by the thread. However, a reader still has to read all posts in all clustered threads to get the insights. To address this shortcoming, \newcite{P18-1062} proposed a system that takes an interleaved text as input and provides the reader with its summaries. Their system is an unsupervised two-step system, first, a conversation disentanglement component clusters the posts thread-wise, and second, a multi-sentence compression component compresses the thread-wise posts to single-sentence summaries. However, this system has two major disadvantages: first, the disentanglement obtained through either supervised \cite{N18-1164} or unsupervised \cite{wang2009context} methods propagate its errors to the downstream summarization task, and therefore, degrades the overall performance, and second, the compression component is restricted to formulate summaries out of disentangled threads, and therefore, cannot bring new words to improve the fluency. 
We aim to tackle these issues % of error propagation 
through an end-to-end trainable encoder-decoder system that takes a variable length input, e.g., interleaved texts, processes it and generates a variable length output, e.g., a multi-sentence summary. %; see the foreground at the bottom in the \figref{interlv_summ}. 
An end-to-end system eliminates the disentanglement component, and thus, the error propagation. Furthermore, the corpus-level vocabulary of the decoder provided it with greater selection of words, and thus, a possibility to improve language fluency. 

In the domain of text summarization, hierarchical encoder, encoding words in a sentence (post) followed by the encoding of sentences in a document (channel), is a very commonly used method \cite{DBLP:conf/conll/NallapatiZSGX16,P18-1013}. However, hierarchical decoding is rare, as many works in the domain aim to comprehend an important fact from single or multiple documents. Summarizing interleaved texts provides us a unique opportunity to employ hierarchical decoding as such texts comprise several facts from several threads. %Thus, hierarchical decoder, decoding of thread representations followed by decoding of words in a thread-summary. 
We also propose novel hierarchical attention, which assists the decoder in its summary generation process with 3-levels of information from the interleaved text; posts, phrases, and words, rather than traditional two levels; post and word \cite{AAAI1714636,DBLP:conf/conll/NallapatiZSGX16,tan2017neural,cheng2016neural}. 

%The attention mechanism is trained end-to-end with the encoder-decoder system, unlike hierarchical attentions that trained through auxiliary dataset obtained by means of heuristics, e.g., \cite{P18-1013}. 

As labeling of interleaved texts is a difficult and expensive task%\cite{barker2016sensei,aker2016automatic,verberne2018creating}
, we devised an algorithm that synthesizes interleaved text-summary pairs corpora of different difficulty levels (in terms of entanglement) from a regular document-summary pairs corpus. Using these corpora, we show the encoder-decoder system not only obviates disentanglement component, but also enhances performance. Further, our hierarchical encoder-decoder system consistently outperforms traditional sequential ones.

To summarize, our contributions are:% are fourfold:
\begin{itemize}
\item We propose an end-to-end encoder-decoder system over pipeline to obtain a quick overview of interleaved texts.     
%\item To the best of our knowledge, we are the first to 
\item To the best of our knowledge, we are first to use a hierarchical decoder to obtain multi-sentence abstractive summaries from texts.
\item We propose a novel hierarchical attention that integrates information from 3 levels; posts, phrases and words, and is trained end-to-end.
\item We devise an algorithm that synthesizes interleaved text-summary corpora, on which we verify pipeline system vs. encoder-decoder, sequential vs. hierarchical decoding, 2- vs. 3-level hierarchical attention. Overall, the proposed system attains 20-40\% performance gains on both real-world (AMI) and synthetic datasets. 
\end{itemize} 
\section{Related Work}
%Quite often multi-party conversations, e.g. news comments, social media conversation and activity report, have tens of posts discussing several different matters pertaining to a subject.
\newcite{ma2012topic,aker2016automatic,P18-1062} designed earlier systems that summarize posts in multi-party conversations in order to provide readers with overview on the discussed matters. They broadly follow the same approach: cluster the posts and then extract a summary from each cluster.
 
There are two kinds of summarization: abstractive and extractive. In abstractive summarization, the model utilizes a corpus level vocabulary and generates novel sentences as the summary, while extractive models extract or rearrange the source words as the summary. Abstractive models based on neural sequence-to-sequence (seq2seq) \cite{DBLP:conf/emnlp/RushCW15} proved to generate summaries with higher ROUGE scores than the feature-based abstractive models. Integration of attention into seq2seq \cite{DBLP:journals/corr/BahdanauCB14} led to further advancement of abstractive summarization \cite{DBLP:conf/conll/NallapatiZSGX16,DBLP:conf/naacl/ChopraAR16}.

%There are many possible patterns of organization of the information in texts, e.g., chronological pattern. News articles have an inverted pyramid pattern, i.e., core information in the lead sentences and the extra information in later sentences. A seq2seq model is appropriate for summarization of a news article as it encodes and decodes sequentially. However, in interleaved texts, related information maybe separated; thus a seq2seq model may be competent but not sufficient.

\newcite{P15-1107} proposed an encoder-decoder (auto-encoder) model that utilizes a hierarchy of networks: word-to-word followed by sentence-to-sentence. Their model is better at capturing the underlying structure than a vanilla sequential encoder-decoder model (seq2seq). \newcite{krause2016paragraphs,P18-1240} showed multi-sentence captioning of an image through %a decoder based on a
hierarchical Recurrent Neural Network (RNN), topic-to-topic followed by word-to-word, is better than seq2seq. 
\begin{figure*}[t!]
\centering
\includegraphics[width=0.99\textwidth]{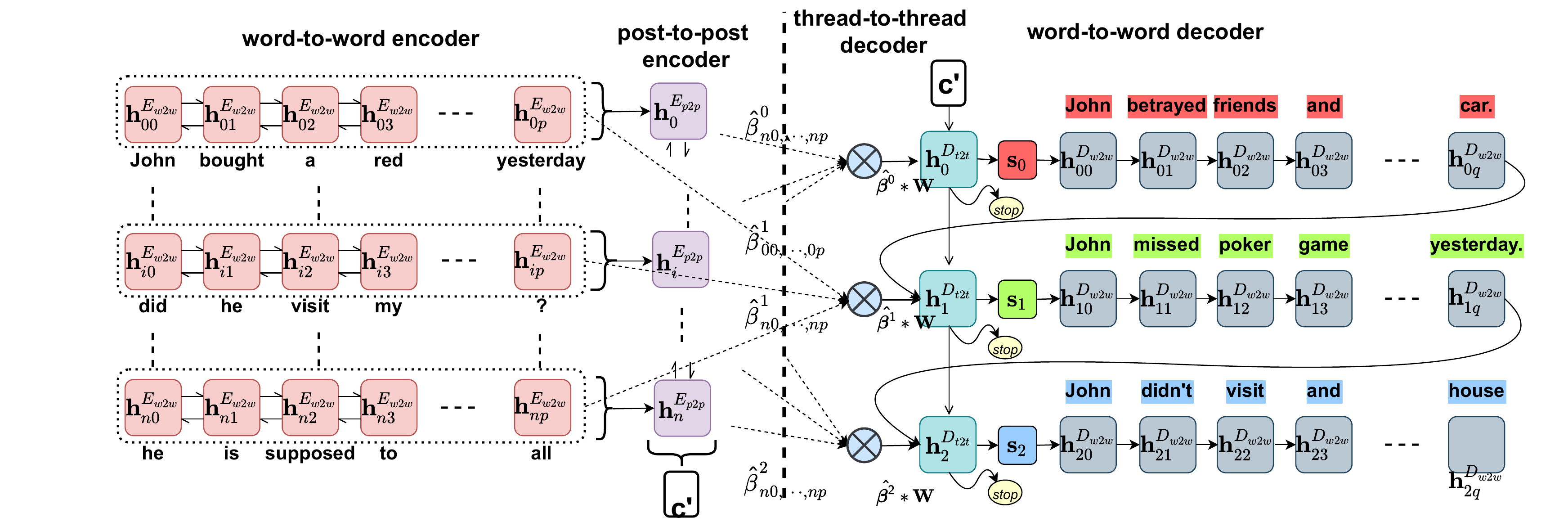}
\caption{Our hierarchical encoder-decoder architecture. On the left, interleaved posts are encoded hierarchically, i.e., word-to-word ($E_{w2w}$) followed by post-to-post ($E_{p2p}$). On the right, summaries are generated hierarchically, thread-to-thread ($D_{t2t}$) followed by word-to-word ($D_{t2t}$). %$\boldsymbol{\hat{\beta}}^k$ are phrase-level attentions and $\mathbf{W}$ are $E_{w2w}$ outputs.
}\figlabel{architecture_joint}
\end{figure*}

These works suggest a hierarchical encoder, with word-to-word encoding followed by post-to-post, will better recognize the dispersed information in interleaved texts. Similarly, a hierarchical decoder, thread-to-thread followed by word-to-word, will intrinsically disentangle the posts, and therefore, generate more appropriate summaries.

\newcite{DBLP:conf/conll/NallapatiZSGX16} devised a hierarchical attention mechanism for a seq2seq model, where two levels of attention distributions over the source, i.e., sentence and word, are computed at every step of the word decoding. Based on the sentence attentions, the word attentions are rescaled. \newcite{P18-1013} slightly simplified this mechanism and computed the sentence attention only at the first step. Our hierarchical attention is more intuitive and computes new sentence attentions for every new summary sentence, and unlike \newcite{P18-1013}, is trained end-to-end.

\section{Model}
\subsection*{Problem Statement}
We aim to design a system that when given a sequence of posts, $\mathit{C} = \langle\mathit{P}_1,\ldots,\mathit{P}_{|\mathit{C}|}\rangle$, produces a sequence of summaries, $\mathit{T} = \langle\mathit{S}_1,\ldots,\mathit{S}_{|\mathit{T}|}\rangle$. For simplicity and clarity, unless otherwise noted, we will use lowercase italics for variables, uppercase italics for sequences, lowercase bold for vectors and uppercase bold for matrices.

\subsection{Encoder}
The hierarchical encoder (see \figref{architecture_joint} left hand section) is based on \newcite{AAAI1714636}, where word-to-word and post-to-post encoders are bi-directional LSTMs. %We refer to \cite{AAAI1714636} for further details.
The word-to-word BiLSTM encoder ($E_{w2w}$) runs over word embeddings of post $\mathit{P}_i$ and generates a set of hidden representations, $\langle\mathbf{h}^{{E_{w2w}}}_{i,0},\ldots,\mathbf{h}^{{E_{w2w}}}_{i,p}\rangle$, of $d$ dimensions. 
The average pooled value of the word-to-word representations of post $\mathit{P}_i$ ($\frac{1}{p}\sum_{j=0}^{p} \mathbf{h}^{{E_{w2w}}}_{i,j}$) is input to the post-to-post BiLSTM encoder ($E_{t2t}$), which then generates a set of representations, $\langle\mathbf{h}^{E_{p2p}}_{0},\ldots,\mathbf{h}^{E_{p2p}}_{n}\rangle$, corresponding to the posts. 
Overall, for a given channel $\mathit{C}$, output representations of word-to-word, $\mathbf{W}$, and post-to-post, $\mathbf{P}$, has $n\times p\times 2d$ and $n\times 2d$ dimensions respectively.

\subsection{Decoder}
%Two major decoding approaches have been utilized in multi-sentence image captioning. \cite{P18-1240} and \cite{krause2016paragraphs} use a top-level RNN, such as thread-to-thread in our case, that takes an image representation and computes topic representations, and a low-level RNN, such as word-to-word in our case, that takes those topic representations and generates sentences corresponding to them. However, a major issue with their systems is the repetition of some of the sentences in a multi-sentence caption. To address this issue, \cite{DBLP:conf/miccai/XueXLXATH18} proposed a top-level network instead of an RNN, that at each step not only takes an image but also an encoded representation of the previously generated caption sentence and then computes a topic representation. In summary, \cite{P18-1240} and \cite{krause2016paragraphs} use a top-level decoder which keeps track of its regions to generate a new topics, while \cite{DBLP:conf/miccai/XueXLXATH18} uses a top-level decoder which takes feedback from the low-level decoder to generate a new topic. Further, \cite{DBLP:conf/miccai/XueXLXATH18} approach is intuitively similar to \cite{P15-1107} hierarchical decoding. 

Our hierarchical decoder structure and arrangement is similar to \newcite{P15-1107} hierarchical auto encoder, with two uni-directional LSTM decoders, thread-to-thread and word-to-word (see right-hand side in \figref{architecture_joint}), however, in terms of inputs, initial states and attentions it differs a lot, which we explain in the next two sections.

The initial state $\mathbf{h}^{D_{t2t}}_{0}$ of the thread-to-thread LSTM decoder ($f^{D_{t2t}}$) is set with a feedforward-mapped representation of an average pooled post representations ($\mathbf{c}^\prime = \frac{1}{n}\sum_{i=0}^{n} \mathbf{h}^{p2p}_{i}$). 
At each step $k$ of the $f^{D_{t2t}}$, a sequence of attention weights, $\langle\mathit{\hat{\beta}}^{k}_{0,0},\ldots,\mathit{\hat{\beta}}^{k}_{n,p}\rangle$, corresponding to the set of encoded word representations, $\langle\mathbf{h}^{w2w}_{0,0},\ldots,\mathbf{h}^{w2w}_{n,p}\rangle$ are computed utilizing the previous state, $\mathbf{h}^{D_{t2t}}_{k-1}$. We will elaborate the attention computation in the next section.

A weighted representation of the words (crossed blue circle) is then computed: $\sum_{i=1}^{n}\sum_{j=1}^{p}\hat{\beta}^{k}_{i,j}\mathbf{W}_{ij}$, 
Additionally, we use the last hidden state $\mathbf{h}^{D_{w2w}}_{k-1,q}$ of the word-to-word decoder LSTM (${D_{w2w}}$) of the previously generated summary sentence as the second input to compute the next state of thread-to-thread decoder, i.e., $\mathbf{h}^{D_{t2t}}_{k}$. The motivation is to provide information about the previous sentence.

The current state $\mathbf{h}^{D_{t2t}}_{k}$ is passed through a single layer feedforward network and a distribution over STOP=1 and CONTINUE=0 is computed: 
\begin{equation}
\mathit{p}_{k}^{STOP} = \sigma(\mbox{g}\left({\mathbf{h}^{D_{t2t}}_{k}}\right)) 
\eqlabel{stop_predict}
\end{equation} where $\mbox{g}$ is a feedforward network. In \figref{architecture_joint}, the process is depicted by a yellow circle. The thread-to-thread decoder keeps decoding until  $\mathit{p}_{k}^{STOP}$ is greater than 0.5.

Additionally, the current state $\mathbf{h}^{D_{t2t}}_{k}$ and inputs to $D_{t2t}$ at that step are passed through a two-layer feedforward network \text{r} followed by a dropout layer %$\tanh$ activation 
to compute the thread representation $\mathbf{s}_k = %\tanh(
\mbox{r}\left({\mathbf{h}^{D_{t2t}}_{k};\mathbf{h}^{D_{w2w}}_{k-1,q};\boldsymbol{\hat{\beta}}^k*\mathbf{W}}\right)$.
%\begin{equation}
%\mathbf{s}_i = tanh(\mbox{k}\left({\mathbf{h}^{D_{t2t}}_{i}}\right))
%\end{equation}

Given a thread representation $\mathbf{s}_k$, the word-to-word decoder generates a summary for the thread. Our word-to-word decoder is based on \newcite{DBLP:journals/corr/BahdanauCB14}. It is a unidirectional attentional LSTM ($f^{D_{w2w}}$); see the right-hand side of \figref{architecture_joint}. %, which uses a thread representation as an input and generates a summary. 
We refer to \cite{DBLP:journals/corr/BahdanauCB14} for further details.%In \figref{architecture_topic_dec} the right side depicts the process.

\subsection{Hierarchical Attention}
\begin{figure}[ht!]
\centering
\includegraphics[width=0.5\textwidth]{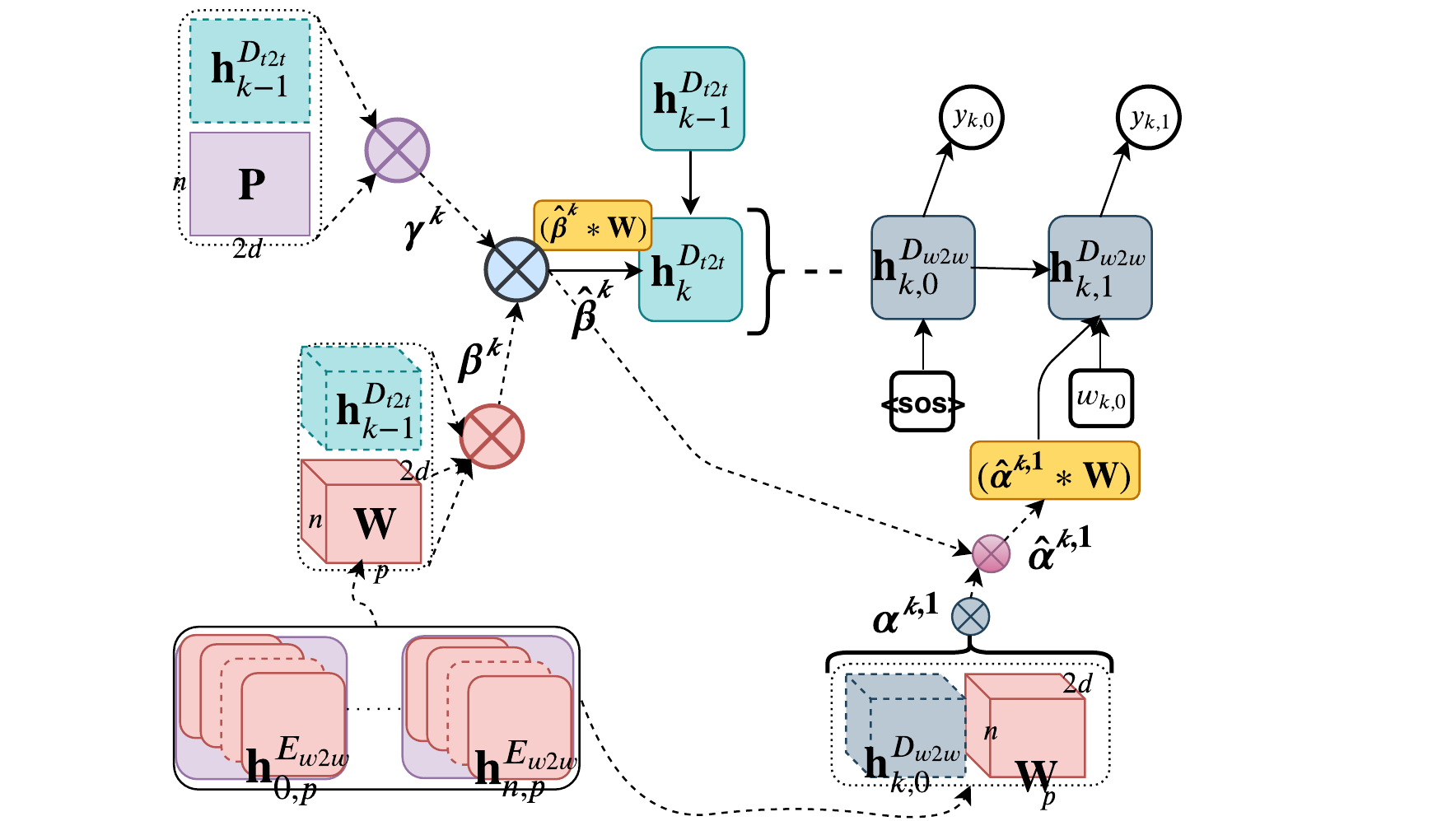}
\caption{Hierarchical attention mechanism. Dotted lines indicate involvement in the mechanism. %There are three successive levels of attentions, i.e, from posts ($\gamma$) to phrases ($\beta$) to words ($\alpha$).
}\figlabel{architecture_hierattn}
\end{figure}

\begin{table*}[t!]
\begin{center}
\resizebox{0.99\textwidth}{!}{
\footnotesize
%\begin{tabular}{p{0.95\linewidth}}%{lp{0.92\linewidth}}
\begin{tabular}{p{0.60\textwidth}|p{0.35\textwidth}}
\hline
\ding{51} this study was conducted to evaluate the influence of e\ldots&\multirow{4}{5cm}{\ding{51} caffeine in sport . influence of endurance exercise on the urinary caffeine concentration .}\\
\ding{70} to assess the effect of a program of supervised fitness\ldots&\\
\ding{70} an 8-week randomized , controlled trial .\ldots&\\
\ding{51} nine endurance-trained athletes participated in a randomised\ldots&\\
%\vdots & \multicolumn{1}{c}{\vdots}\\
\multicolumn{1}{c|}{\ldots}&\multirow{3}{5.5cm}{\ding{70} supervised fitness walking in patients with osteoarthritis of the knee . a randomized , controlled trial .}\\
\ding{81} we examined the effects of intensity of training on ratings\ldots&\\
\ding{81} subjects were recruited as sedentary controls or were randomly\ldots&\\
\ding{81} the at lt group trained at velocity lt and the greater than\ldots&\multirow{2}{5cm}{\ding{81} the effect of training intensity on ratings of perceived exertion .}\\
\ding{51} data were obtained on 47 of 51 intervention patients and 45\ldots&\\
\hline
\end{tabular}
}
\end{center}
\caption{\tablabel{example_dataset1}The left rows contain interleaving of 3 articles with 2 to 5 sentences and the right rows contain their interleaved titles. Associated sentences and titles are depicted by similar symbols.}
\end{table*}

%Our novel dynamic hierarchical attention is based on two popular concepts: attention scores should integrate in top-down manner (\cite{DBLP:conf/conll/NallapatiZSGX16}) and two distributions over same words must be modeled together (\cite{P17-1099}), not just in isolation. %We draw inspiration from \cite{P17-1099} wherein a model learns to merge attention distribution over source tokens with the attention distribution over vocabulary tokens using switch probabilities, and thereby, determines which tokens to copy and which to pull from vocabulary in the process of generating a summary.
%will not be associated to any local feature, leading to many all-zero residuals for the region. For visual words that correspond to visual patterns observed in only a small number of regions, this will lead to substantially downweighted residuals. 

%pointer mechanism and . Basically, in an end-to-end learning, if there are two distributions that involves a common set of tokens then they must be modeled together through a weight parameter, e.g., switch probabilities, but not in isolation. We also model two distributions, but in contrast to See et al., we use them to capture larger-scale and local context, i.e., high-level distribution (beta) and low-level distribution (alpha) over the same source tokens and we model them together by scaling one (alpha) by the other (beta). 

Our novel hierarchical attention works at 3 levels, the post level (corresponding to posts), i.e., $\boldsymbol{\gamma}$, and phrase level (corresponding to source tokens), i.e., $\boldsymbol{\beta}$, and are computed while obtaining a thread representation, $\mathbf{s}$. The word level attention (also corresponding to source tokens), i.e., $\boldsymbol{\alpha}$, is computed while generating a word, $y$, of a summary, $\mathit{S}$.; see \figref{architecture_hierattn}.

We draw inspiration for the hierarchical attention from some of the recent works in computer vision \cite{noh2017large,teichmann2019detect}, in which, they show a convolutional neural network (CNN)-based local descriptor with attention is better at obtaining key points from an image than CNN-based global descriptor. Phrases from posts of interleaved texts are equivalent to visual patterns in images, and thus, extracting phrases is more relevant for thread recognition than extracting posts. Thus, contrary to popular hierarchical attention \cite{DBLP:conf/conll/NallapatiZSGX16,cheng2016neural,tan2017neural}, we have additional phrase-level attention focusing again on words, but with a different responsibility. Further, the popularly held intuition of hierarchical attention, i.e., sentence attention scales word attention, is still intact as gamma (post-attention) scales beta.

At step $k$ of thread decoding, we compute elements of post-level attention, i.e., $\boldsymbol{\gamma}^{k,\cdot}$ as.
\begin{equation}
\gamma^{k}_{i} = \sigma(\mbox{attn}^{\gamma}(\mathbf{h}^{D_{t2t}}_{k-1}, \mathbf{P}_{i})\quad i \in \{1,\dotsc,n\}
\label{eqn:gamma_attn}
\end{equation}, where $\mbox{attn}^{\gamma}$ aligns the current thread decoder state vector $\mathbf{h}^{D_{t2t}}_{i-1}$ to vectors of matrix $\mathbf{P}_{i}$ and then maps aligned vectors to scalar values through a feed-forward network. 
At the same step, we also compute elements of phrase-level attention, i.e, $\boldsymbol{\beta}^{k}_{i,j}$ as.
\begin{equation}
\begin{aligned}
\beta^{k}_{i,j} &= \sigma(\mbox{attn}^{\beta}(\mathbf{h}^{D_{t2t}}_{k-1}, \mathbf{a}_{i,j})) \\ 
& \text{where} \enskip \mathbf{a}_{i,j} = add(\mathbf{W}_{i,j}, \mathbf{P}_{i}),\\
& i \in \{1,\dotsc,n\},\quad j \in \{1,\dotsc,p\}
\label{eqn:beta_attn}
\end{aligned}
\end{equation}, 
%$\mathbf{W}$ in \eqref{beta_align} is a word-to-word encoder representations of dimension $n\times p \times 2l$, 
${add}$ aligns a post representation to its constituting word representations and does element-wise addition, and $\mbox{attn}^{\beta}$ is a feedforward network that maps the current thread decoder state $\mathbf{h}^{D_{t2t}}_{k-1}$ and vector $\mathbf{a}_{i,j}$ to a scalar value.
Importantly, $\sigma(\cdot)$ in $\gamma$ and $\beta$ will allow a thread not to be associated with any relevant phrase, and thereby, indicating a halt in decoding. 

Then, we use $\boldsymbol{\gamma}^{k}$ to rescale phrase-level attentions, $\boldsymbol{\beta}^{k}$ as $\hat{\beta}^{k}_{i,j} = \beta^{k}_{i,j}*\gamma^{k}_{i}$. %below.

%\begin{equation}
%\begin{aligned}
%\hat{\beta}^{k}_{i,j} &= \beta^{k}_{i,j}*\gamma^{k}_{i}\\ 
%& i \in \{1,\dotsc,n\}, \quad j \in \{1,\dotsc,p\}
%\end{aligned}
%\label{eqn:beta_recale}
%\end{equation} 
At step $l$ of word-to-word decoding of summary thread $k$, we compute elements of word level attention, i.e., $\boldsymbol{\alpha}^{k,l}_{i,\cdot}$ as below. 
\begin{equation}
\begin{aligned}
\alpha^{k,l}_{i,j} &= \frac{\exp(\mathbf{e}^{k,l}_{i,j})}{\sum_{i=1}^{n}\sum_{j=1}^{p}\exp(\mathbf{e}^{k,l}_{i,j})}\\
&\text{where} \enskip
\mathbf{e}^{k,l}_{ij}=\mbox{attn}^{\alpha}(\mathbf{h}^{D_{w2w}}_{k,l-1}, \mathbf{a}_{i,j})
\end{aligned}
\label{eqn:alpha_attn}
\end{equation}, and $\mathbf{a}_{k}$ is same as in \eqref{beta_attn} and $\mbox{attn}^{\alpha}$ is a feedforward network that maps the current word decoder state $\mathbf{h}^{D_{w2w}}_{k,l-1}$ and vector $\mathbf{a}_{i,j}$ to a scalar value.

Finally, we use rescaled phrase-level word attentions, $\hat{\boldsymbol{\beta}^{k}}$, for rescaling word level attention, $\alpha^{k,l}$ as $\hat{\alpha}^{k,l}_{i,j} = \hat{\beta}^{k}_{i,j} \times \alpha^{k,l}_{ij}$

%below:
%\begin{equation}
%\hat{\alpha}^{k,l}_{ij} = \frac{\hat{\beta}^{k}_{i,j} \times \alpha^{k,l}_{ij} %\label{eqn:word_attn_wts_s2h}
%\end{equation}

%Overall, top attention, i.e., post-level scales mid one , i.e., high-level, which then scales bottom one, i.e., low-level. 
%\tabref{hattn_comp} compares our hierarchical attention with previous work on hierarchical attention in the text summarization domain.

%\def\evalsepL{0.1cm}
%\def\evalsepS{0.055cm}
%\begin{table}[h!]
%\begin{center}
%\footnotesize
%\caption{\tablabel{hattn_comp} Forms of hierarchical attention. E2E= End2End, i.e., No training labels, $\times words$=for every word in the target, $\times summaries$=for every summary in the target, $\boldsymbol{\gamma}$=sentence(post)-level, $\boldsymbol{\beta}$=high-level} %HierDcode=Hierarchical Decoding}
%\begin{tabular}{p{2.9cm}|c@{\hspace{\evalsepS}}c@{\hspace{\evalsepS}}c@{\hspace{\evalsepL}}c@{\hspace{\evalsepS}}}
%Model &  E2E & Decode & $\boldsymbol{\gamma}$ &  $\boldsymbol{\beta}$\\ \hline
%\cite{DBLP:conf/conll/NallapatiZSGX16}, \cite{cheng2016neural}, \cite{tan2017neural} & yes& seq& $\times words$ & no\\
%\cite{P18-1013} & no & seq & 1 & no\\
%%\cite{P15-1107} & yes & yes & no\\
%Ours & yes & hier & $\times threads$ & yes\\
%\end{tabular}
%\end{center}
%\end{table}

\subsection{Training Objective}
We train our hierarchical encoder-decoder network similarly to an attentive seq2seq model \cite{DBLP:journals/corr/BahdanauCB14}, but with an additional weighted sum of sigmoid cross-entropy loss on stopping
distribution; see \eqref{stop_predict}. Given a thread summary, $\mathit{Y}^k = \langle \mathit{w}^{k,0},\ldots, \mathit{w}^{k,q}\rangle$, our word-to-word decoder generates a target $\hat{\mathit{Y}}^k = \langle \mathit{y}^{k,0},\ldots, \mathit{y}^{k,q} \rangle$, 
with words from a same vocabulary $\mathit{U}$. %Irrespective of top-level network in the hierarchical decoder, 
We  train our model end-to-end by minimizing the objective 
given in \eqref{log_likelihood}. 
\begin{equation}
\begin{aligned}
{\underset{k=1}{\overset{m}{\sum}}}{\underset{l=1}{\overset{q}{\sum}}}&\log{p}_\theta\left(y^{k,l}\vert\textit{w}_{k,\cdot<l}, \textbf{W}\right)
\eqlabel{log_likelihood}\\ 
&+\lambda  {\underset{k=1}{\overset{m}{\sum}}}\textit{y}_{k}^{STOP}\log(p_{k}^{STOP})
\end{aligned}
\end{equation}
%where $\mathbf{C}$ is word representations matrix.

\section{Dataset}
%A large number of conversations occur every day in social media, e.g., Reddit, Slack and Twitter; however, summaries of those conversations are not available. Therefore, 
Obtaining labeled training data for conversation summarization is challenging. The available ones are either extractive \cite{verberne2018creating} or too small \cite{barker2016sensei,anguera2012speaker} to train a neural model. To get around this issue and thoroughly verify the proposed architecture, we synthesized a dataset by utilizing a corpus of conventional texts for which summaries are available. %The selection of the texts is restricted to those that have their sentences arranged in a structure that is comparable to the arrangement of posts in a thread. 
%The PubMed corpus contains a type of article, i.e., randomized controlled trials (RCT), where sentences in the abstract are structured into sections and the title of the article is an abstractive summary of the information from these sections (\cite{I17-2052}). Further, Stack Exchange is a question-and-answer site for several diverse fields topics. In this dataset, every question (with multiple sentences) has a title, and zero, one or many answers. The title forms the gist or abstract of the question. 
We create two corpora of interleaved texts: one from the abstracts and titles of articles from the PubMed corpus% of randomized controlled trials (RCT) 
 and one from the questions and titles of Stack Exchange questions.
A random interleaving of sentences from a few PubMed abstracts or Stack Exchange questions roughly resembles interleaved texts, and correspondingly interleaving of titles resembles its multi-sentence summary. %We devised an algorithm for creating the synthetic interleaved texts; see \algref{interleave_algo}. 

The algorithm that we devised for creating synthetic interleaved texts is defined in detail in the Appendix. %Given a set of abstracts and titles and a range for , a set of concatenated interleaved texts and summaries are returned. 
The number of abstracts to include in the interleaved texts is given as a range (from $a$ to $b$) and the number of sentences per abstract to include is given as a second range (from $m$ to $n$). We vary the %$\textsc{Interleave}$ 
parameters as below and create three different corpora for experiments: \textbf{Easy} ($a$=2, $b$=2, $m$=5 and $n$=5), \textbf{Medium} ($a$=2, $b$=3, $m$=2 and $n$=5) and \textbf{Hard} ($a$=2, $b$=5, $m$=2 and $n$=5).
%\begin{itemize}
%\item Easy: $a$=2, $b$=2, $m$=5 and $n$=5
%\item Medium: $a$=2, $b$=3, $m$=2 and $n$=5 
%\item Hard: $a$=2, $b$=5, $m$=2 and $n$=5 
%\end{itemize}
\tabref{example_dataset1} shows an example of a data instance in the Hard Interleaved RCT corpus. 
%RCT articles have MeSH descriptors\footnote{https://meshb.nlm.nih.gov/treeView} that categorizes them into 16 categories. We use this information to split the above corpora into Train, Validation and Test. Training uses 14 categories and one each is used for test and validation. We didn't do any hyper-parameter tuning; however, we use validation for early stopping. %to select the best performing model.
\def\evalsep{0.20cm}
\def\perlevelsep{0.125cm}
\def\perlevelsepBase{0.08cm}
\begin{table*}[t!]
\begin{center}
\footnotesize
\resizebox{1.0\linewidth}{!}{
\begin{tabular}
{l@{\hspace{\perlevelsepBase}}|c|
c@{\hspace{\perlevelsep}}c@{\hspace{\perlevelsep}}c|c@{\hspace{\perlevelsep}}c@{\hspace{\perlevelsep}}c|c@{\hspace{\perlevelsep}}c@{\hspace{\perlevelsep}}c|}
\multirow{2}{1cm}{Input Text} &\multirow{2}*{Model} &\multicolumn{3}{c|}{\bf Easy} &\multicolumn{3}{c|}{\bf Medium}&\multicolumn{3}{c|}{\bf Hard}\\
& & Rouge-1 &  Rouge-2 &  Rouge-L & Rouge-1 &  Rouge-2 &  Rouge-L& Rouge-1 &  Rouge-2 &  Rouge-L\\ 
\hline
\tikzmark{top left 1}ind&  seq2seq & 35.09& 28.72& 13.16& 36.31& 28.78& 13.45& 37.74& 28.72& 13.76\\
dis& seq2seq& \bf36.38& \bf29.90& \bf14.78& 35.63&	28.45&	13.98&	37.87&	28.85&	14.77\\
dis& hier2hier & 35.30&	28.93&	13.35&	\bf37.30& \bf29.83& \bf14.90& \bf39.09& \bf30.11& \bf15.22\\
\hline
kmn& seq2seq(dis) & 34.48& 27.51& 13.31& 34.05&	26.58&	13.14&	35.54&	26.36&	13.65\tikzmark{bottom right 1}\\
%\cdashline{1-11}
\tikzmark{top left 2}kmn& seq2seq & 34.28& 27.84& 13.86& 34.89& 27.42& 13.68& 31.22& 23.37& 11.77\\
\hline
kmn& compress & 30.04& 19.83& 10.75& 29.37& 17.54&	10.43& 29.11& 15.76& 10.13\\
ent& seq2seq & 35.78& \bf28.89& \bf14.62&	35.20&	27.44&	13.54& 32.46&	24.17&	12.17\\
ent& hier2hier & \bf35.88& 28.47& 13.33& \bf37.29&	\bf29.63& \bf 14.95& \bf37.11& \bf27.97& \bf14.26\tikzmark{bottom right 2}\\
\end{tabular}
\DrawBox[ thick, draw=blue, dotted]{top left 1}{bottom right 1}
\DrawBox[ thick, draw=green, dashed]{top left 2}{bottom right 2}

}
\end{center}
\caption{\tablabel{base_perf_comp} Summarization performance (Rouge Recall-Scores) comparing models when the threads are disentangled (top blue dotted section, upper bounds) and when the threads are entangled (bottom green dashed section, real-world) on the Easy, Medium and Hard Pubmed Corpora. \textbf{ind} = individual, \textbf{dis} = disentangled (ground-truth), \textbf{kmn} = K-means disentangled and \textbf{ent} = entangled. %Top 4 rows show results when ground-truth clusters of threads are available. Bottom 4 rows show real-world entangled scenarios, among which 
In the middle, the first row shows a seq2seq model trained on ground-truth disentangled texts and tested on unsupervised disentangled texts, and the second row shows a seq2seq model trained and tested on unsupervised disentangled texts. The best performance for the entangled threads and for the disentangled threads are in bold.
%In the bottom, the first row (kmn + compress) is a pipeline system of Shang et al.'s and the bottom two rows are end-to-end.
}
\end{table*}

\section{Experiments}
%\textbf{Evaluation Metrics:} %The DUC short summary-based competition  \footnote{http://duc.nist.gov/duc2004/tasks.html} evaluated all competing systems using ROUGE \cite{lin:2004:ACLsummarization}, which provides scores for different n-gram matches. Several other studies on text-summarization also evaluated their system using ROUGE; \cite{P17-1099,DBLP:conf/conll/NallapatiZSGX16,DBLP:conf/naacl/ChopraAR16,DBLP:conf/emnlp/RushCW15}. Following these works, 
We report ROUGE-1, ROUGE-2, and ROUGE-L as the quantitative evaluation of the models.
%\textbf{Parameters:}
%The Pubmed Medium and Hard corpora train, test and validation have approximately 290k, 6k and 1.5k instances, respectively. The Stack-Exchange Medium corpus train has 180k and Hard has 210k instances, while the both corpora have test and validation of approximately 5k and 4k instances, respectively. The remaining hyper-parameters are described in detail in the Appendix.
The hyper-parameters for experiments are described in detail in the Appendix and remain the same unless otherwise noted.

\subsection{Upper-bound}

In upper-bound experiments, we check the impact of disentanglement on the abstractive summarization models, e.g., seq2seq and hier2hier. In order to do this, firstly, we provide the ground-truth disentanglement (cluster) information and evaluate the performance of these models. Secondly, we let the models to perform either end-to-end or two-step summarization. In order to perform these experiments, we compiled three corpora of different entanglement difficulty using Pubmed corpus of MeSH type Disease and Chemical\footnote{interleaving is performed within a MeSH type}. The training, evaluation and test sets are of sizes of 170k, 4k and 4k respectively. 

The seq2seq model can use ground-truth disentanglement information in two ways, i.e., summarize threads individually or summarize concatenated threads. The first two rows in \tabref{base_perf_comp} compares performance of those two sets of experiments. Clearly, seq2seq model can easily detect thread boundary in concatenated threads and perform as good as individual model. However, hier2hier is better than seq2seq in detecting thread boundaries as indicated by its performance gain on Medium and Hard corpora (see row 3 in \tabref{base_perf_comp}), and therefore, sets the upper bound for interleaved text summarization.  

Additionally, we also utilize \newcite{P18-1062}'s unsupervised disentanglement component and cluster the entangled threads. Importantly, their disentanglement component requires a fixed cluster size as an input; however, our Medium and Hard corpora have a varying cluster size, and therefore, we give their system benefit of the doubt and input the maximum cluster size, i.e., 3 and 5 respectively. We sort the clusters by their association to a sequence of summary, where the association is measured by Rouge-L between them. We then take the seq2seq trained on ground-truth disentanglement and test it on these unsupervised-disentangled texts to understand the strength of unsupervised clustering. The performance of the pre-trained model remains somewhat similar (see row 2 and 4), indicating a strong disentanglement component. We also train and test a seq2seq on unsupervised-disentangled texts; however, its performance lowers slightly (see row 5), which we believe is due to noise inserted by heuristic sorting of clusters.

In real-world scenarios, i.e., without ground-truth disentanglement, \cite{P18-1062}'s unsupervised two-step system performs worse than seq2seq on unsupervised disentanglement (see row 5 and 6), the reason being a seq2seq model trained on a sufficiently large dataset is better at summarization than the unsupervised sentence compression (extractive) method. At the same time, a seq2seq model trained on entangled texts performs similar to a seq2seq trained on unsupervised disentangled texts (see row 5 and 7), and thereby, showing that the disentanglement component is not necessary. Finally, a hier2hier trained on entangled texts is the only model that reaches closest to the upper-bound set by hier2hier on disentangled texts (see row 3 and 8).

%it on the Easy corpus. We also ran \cite{P18-1062}'s unsupervised two-step system on the test set of the Easy Interleaved Pubmed corpus. %As the \cite{P18-1062} system is unsupervised, it doesn't need training.
%Additionally, we also utilized \cite{P18-1062}'s clustering component to first cluster the interleaved texts of the corpus, and then the disentangled corpus is used to train the seq2seq model. We refer to the latter as cluster$\rightarrow$seq2seq. The performance comparison of \cite{P18-1062} and the two seq2seq models are shown in \tabref{base_perf_comp}. Clearly, seq2seq performs better than \cite{P18-1062}, the reason being a seq2seq model trained on a sufficiently large dataset is better at summarization than the unsupervised sentence compression (extractive) method. The lower performance of cluster$\rightarrow$seq2seq in comparison to seq2seq shows that a disentanglement component is unnecessary in easy scenario.% but also illustrates the error propagation of disentanglement to summarization. 
%Furthermore, we utilize the ground-truth labels and disentangle the interleaved posts of a sampled (150k) Hard Corpus, and re-run the seq2seq model. The results (see \tabref{base_perf_comp}) show seq2seq (row 3) takes advantage of cluster labels but hier2hier (row 4) does more than it. 

\begin{table*}[t!]
\begin{center}
\footnotesize
\begin{tabular}
{l|l|
c@{\hspace{\perlevelsep}}c@{\hspace{\perlevelsep}}c|c@{\hspace{\perlevelsep}}c@{\hspace{\perlevelsep}}c|}
Corpus& \multirow{2}*{Model} &\multicolumn{3}{c|}{\bf Pubmed} &\multicolumn{3}{c|}{\bf Stack Exchange}\\
Difficulty &  & Rouge-1 &  Rouge-2 &  Rouge-L & Rouge-1 &  Rouge-2 &  Rouge-L\\ 
\hline
\multirow{2}*{Medium} & seq2seq &  30.67 & 11.71 & 23.80  &  18.78 & 03.52 & 14.73\\
& hier2hier & \bf32.78 & \bf12.36 & \bf25.33 & \bf24.34 & \bf05.07 & \bf18.63\\
\hline
\multirow{2}*{Hard} & seq2seq &  29.07 & 10.96 & 21.76 &  20.21 & 04.03 & 14.93\\
& hier2hier & \bf 33.36 & \bf 12.69 & \bf 24.72 & \bf 24.96 & \bf 05.56 & \bf 17.95\\
\end{tabular}
\end{center}
\caption{\tablabel{hier_perf_comp} Rouge Recall-Scores on the Medium and Hard Corpora. The base Pubmed has abstract-summary pairs of 10 MeSH types, while base Stack Exchange has posts-question pairs from 12 topics.}
\end{table*}

\section{Seq2seq vs. hier2hier models}
Further, we compare the proposed hierarchical approach against the seq-to-seq approach in summarizing the interleaved texts by experimenting on the Medium and Hard corpora obtained from much-varied base document-summary pairs. We interleave Pubmed corpus of 10 MeSH types, e.g., anatomy and organism. Similarly, we interleave Stack Exchange posts-question pairs of 12 different categories with regular vocabularies, e.g., science fiction and travel. As before, the interleaving is performed within a type or category. The training, evaluation and test sets of Pubmed are of sizes 280k, 5k and 5k and Stack Exchange are of sizes 140k, 4k and 4k respectively. Results in \tabref{hier_perf_comp} shows that a noticeable improvement is observed on changing the decoder to hierarchical, i.e., 1.5-3 Rouge points in Pubmed and 2-4.5 points in Stack Exchange.
%an increase in the complexity of interleaving scantly impacts the performances. Though, in case of Pubmed, seq2seq depreciates by 1-2 points while hier2hier remains nearly same with increase of complexity. In Stack Exchange corpora, as Hard corpora has more training data ($+$30k) seq2seq improves slightly. However, a noticeable improvement is observed on changing the decoder to hierarchical, i.e., 1.5-3 Rouge points in Pubmed and 2-4.5 points in Stack Exchange depending on the interleaving complexity.

Additionally, we evaluated models strength in recognizing threads where summaries are ordered by the location of each thread’s greatest density. Here, density refers to smallest window of posts with over 50\% of posts belonging to a thread; e.g.,
post1-thread1, post1-thread2, post-2-thread2, post2-thread1, post3-thread1, post4-thread1 $\rightarrow$ thread2-summary, thread1-summary. In this example, although thread1 occurs early, as the majority of posts on thread1 occurs latter, therefore, its summary also occurs later. %Further, we also compiled corpora to mimic real world conversation interleaving, wherein the sequence of summaries for interleaved posts may not follow the sequential occurrence of posts, see 17-18 in \algref{interleave_algo}). An example, a post mention an action in the very beginning of conversation, but does only elaborates it at the end. Therefore, summary corresponding to the action should be at the end. %So, we rather compute span covering $>50\%$ of topics, and use these span beginning for ordering the summaries. 
So, we create Medium and Hard corpora of the Stack Exchange with summaries sorted by thread density and perform abstractive summarization studies. As seen in \tabref{hier_dnsty_comp}, both the seq2seq and hier2hier models perform similar to the corpora with summaries sorted by thread occurrence (see \tabref{hier_perf_comp}), which indicates a strong disentanglement in such abstractive models irrespective of summary arrangement. %slightly lower as it is tougher task%compared to the first occurrence results; see \tabref{hier_perf_comp}. 
In addition, the hier2hier model is still consistently better than the seq2seq model. 

\begin{table}[h!]
\footnotesize
\begin{center}
\begin{tabular}
{l|c@{\hspace{\perlevelsep}}c@{\hspace{\perlevelsep}}c|}
&\multicolumn{3}{c|}{\bf Medium Corpus}\\
Model & Rouge-1 &  Rouge-2 &  Rouge-L  \\
\hline
seq2seq & 19.67 & 03.88 & 15.37 \\
hier2hier & \bf23.97 & \bf05.63 & \bf18.75 \\
& \multicolumn{3}{c|}{\bf Hard Corpus}\\
seq2seq &  19.62 & 03.71 & 14.90\\
hier2hier & \bf 24.14 & \bf 05.00 & \bf 17.25\\
\end{tabular}
\end{center}
\caption{\tablabel{hier_dnsty_comp} Rouge Recall-Scores of models on the Stack Exchange Medium and Hard Corpus.}
\end{table}

To understand the impact of hierarchy on the hier2hier model, we perform an ablation study and use the Hard Pubmed corpus for the experiments, and \tabref{hier_ablt_comp} shows the results. Clearly, adding hierarchical decoding already provides a boost in the performance. Hierarchical encoding also adds some improvements to the performance; however, the enhancement attained in training and inference speed by the hierarchical encoding is much more valuable (see Figure 1 in Appendix C)%We will discuss it in depth later. 
\footnote{hier2hier takes $\approx$1.5 days for training on a Tesla V100 GPU, while seq2seq takes $\approx$4.5 days}. Thus, hier2hier model not only achieves greater accuracy but also reduces training and inference time.

\begin{table}[h!]
\footnotesize
\begin{center}
\begin{tabular}
{l|
c@{\hspace{\perlevelsep}}c@{\hspace{\perlevelsep}}c|}
\Tstrut &\multicolumn{3}{c|}{\bf Pubmed Hard Corpus}\\
Model & Rouge-1 &  Rouge-2 &  Rouge-L\\
\hline 
seq2seq &  29.07 & 10.96 & 21.76\\
seq2hier & 32.92  & 11.87 & 24.43\\
hier2seq &  31.86 & 11.9 & 23.57\\
hier2hier & \bf 33.36 & \bf 12.69 & \bf 24.72\\
\end{tabular}
\end{center}
\caption{\tablabel{hier_ablt_comp} Rouge Recall-Scores of ablated models (encoder-decoder) on the Pubmed Hard Corpus.}
\end{table}

%Importantly, hier2hier converges much earlier than the seq2seq and also reaches a lower training loss (\figref{hier_perf_comp2} in Appendix%shows the running average training loss of the seq2seq and hier2hier models on the Stack Exchange Hard corpus.
%). 

\section{Hierarchical attention}
%The contribution of the phrase-level attentions in a hierarchical decoder is two-fold: computing the thread representation and rescaling the word-level attentions. 
To understand the impact of hierarchical attention on the hier2hier model, we perform an ablation study of post-level attentions ($\boldsymbol{\gamma}$) and phrase-level attentions ($\boldsymbol{\beta}$), using the Pubmed Hard corpus.% for the experiments. 

\def\perlevelsepHatn{0.2cm}
\begin{table}[h!]
\begin{center}
%\caption{\tablabel{hier_attn_ablt} Rouge Recall-Scores of ablated models on the Hard Pubmed Corpus.}
\footnotesize
\resizebox{1.0\linewidth}{!}{
\begin{tabular}
{l@{\hspace{\perlevelsepHatn}}|
c@{\hspace{\perlevelsep}}c@{\hspace{\perlevelsep}}c|}
%\Tstrut &\multicolumn{3}{c|}{\bf Hard Pubmed Corpus}\\
Model & Rouge-1 &  Rouge-2 &  Rouge-L\\
\hline 
hier2hier$(+\boldsymbol{\gamma}+\boldsymbol{\beta})$ & \bf 33.36 & \bf 12.69 & \bf 24.72\\
hier2hier$(-\boldsymbol{\gamma}+\boldsymbol{\beta})$ & 32.65  & 12.21 & 24.23\\
hier2hier$(+\boldsymbol{\gamma}-\boldsymbol{\beta})$ &  31.28 & 10.20 & 23.49\\
hier2hier(Li et al.) &  29.83 & 09.80 & 22.17\\
hier2hier$(-\boldsymbol{\gamma}-\boldsymbol{\beta})$ &  30.58 & 10.00 & 22.96\\
seq2seq &  29.07 & 10.96 & 21.76\\
\end{tabular}
}
\end{center}
\caption{\tablabel{hier_attn_ablt} Rouge Recall-Scores of ablated models (attentions) on the Hard Pubmed Corpus.}
\end{table}

\tabref{hier_attn_ablt} shows the performance comparison. $\boldsymbol{\gamma}$ attention improves the performance (0.5-1) of hierarchical decoding but not a lot. The phrase-level attention, i.e., $\boldsymbol{\beta}$ is very important as without it the model performance is noticeably reduced (Rouge values decrease from 2-3). 
The closest hierarchical attentions to ours, i.e., \cite{AAAI1714636,DBLP:conf/conll/NallapatiZSGX16,tan2017neural,cheng2016neural} do not use $\boldsymbol{\beta}$, and therefore, is equivalent to hier2hier$(+\boldsymbol{\gamma}-\boldsymbol{\beta})$, whose performs worse than hier2hier$(-\boldsymbol{\gamma}+\boldsymbol{\beta})$ and hier2hier$(+\boldsymbol{\gamma}+\boldsymbol{\beta})$, thus signifying importance of $\beta$. We also include \newcite{P15-1107} type post-level attention technique in the comparison, where  a softmax $\gamma$ instead of $\sigma(\cdot)$ based $\gamma$ and $\beta$ is used to compute thread representation. Results indicate $\sigma(\cdot)$ fits better in this case. %, and in contrast to us, they do not reuse $\boldsymbol{\beta}$ for re-scaling $\boldsymbol{\alpha}$, and thereby, closest to hier2hier$(+\boldsymbol{\gamma}-\boldsymbol{\beta})$. 
%Evidently, re-utilization gives hier2hier$(+\boldsymbol{\gamma}-\boldsymbol{\beta})$ an edge over (Li, Luong and Jurafsky, 2015). 
Lastly, removing both the $\boldsymbol{\gamma}$ and $\boldsymbol{\beta})$ makes the hier2hier similar to seq2seq, except a few more parameters, i.e., two additional LSTM, and the performance is also very similar. 
%Moreover, the enhancement in case of the hierarchical encoders is higher than the sequential ones as it integrates more appropriately to the hierarchical arrangement of encoder information.

%In threadLSTM, the thread-to-thread decoder, ${D_{t2t}}$, utilizes the post-level attention through its input% and then generates a thread representation $\mathbf{s}_\cdot$
%; see \secref{topic_dec}. Also, depending on encoder type, the word-level attentions in thread\-LSTM are rescaled either using  \eqref{word_attn_wts_h2h} or \eqref{word_attn_wts_s2h}. In this ablation study, we assign 1 to $\boldsymbol{\beta}$ values, and thereby, changing the input of ${D_{t2t}}$ to %computation of the thread-representation to 
%$\frac{1}{n}\sum_{j=1}^{n}\mathbf{m}_{j}$, and also word attention from $\boldsymbol{\hat{\alpha}}$ (\eqref{word_attn_wts_h2h} or \eqref{word_attn_wts_s2h}) to default \cite{DBLP:journals/corr/BahdanauCB14}'s $\boldsymbol{\alpha}$. In feedbackLSTM, the ablation changes remain the same as the threadLSTM except the network $\mbox{k}$ computing thread representation now takes a simple average ($\frac{1}{n}\sum_{j=1}^{n}\mathbf{m}_{j}$) %. and $\mathbf{h}^{D_{t2t}}_{i-1}$ 
%as the input.

%\tabref{hattn_perf_comp} shows the performance comparison. Clearly, models with hierarchical attention have better performance than ones without. Moreover, the enhancement in case of the hierarchical encoders is higher than the sequential ones as it integrates more appropriately to the hierarchical arrangement of encoder information.

\section{AMI Experiments}
We also experimented both abstractive models; seq2seq and hier2hier, on the popular meeting AMI corpus \cite{7529878bc1a143dbad4fa019e742fdb8}, and compare them against \newcite{P18-1062} two-step system. We follow the standard train, eval and test split. Results in \tabref{abst_ami} show hier2hier outperforms both systems by a large margin.
%We find models trained for 50k iteration on Pubmed Hard corpora (a=6, b=10, m=2 and n=3) and then fine-tuned on the popular meeting AMI corpus \cite{7529878bc1a143dbad4fa019e742fdb8} already reaches competitive results (Rouge-1=39.81, Rouge-2=11.35) against SOA system on AMI (\cite{P18-1062} system.  Rouge-1=37.86 and Rouge-2=07.84).

\begin{table}[h!]
\begin{center}
\footnotesize
\resizebox{1.0\linewidth}{!}{
\begin{tabular}
{l@{\hspace{\perlevelsepHatn}}|
c@{\hspace{\perlevelsep}}c@{\hspace{\perlevelsep}}c|}
%\Tstrut &\multicolumn{3}{c|}{\bf Hard Pubmed Corpus}\\
Model & Rouge-1 &  Rouge-2 &  Rouge-L\\
\hline 
Shang et al.& 29.00 & -&-\\
seq2seq & 31.60 & 10.60 & 25.03\\
hier2hier & \bf39.75 & \bf12.75 & \bf25.41\\
\end{tabular}
}
\end{center}
\caption{\tablabel{abst_ami} Rouge F1 Scores of models on AMI Corpus with summary size 150.}
\end{table}
\section{Discussion}
%Occurrence of interleaved texts is common; however, readers often don't have the patience to wade through them. Surveys have shown that summarized contents are easy to consume, and thus, many organizations have begun displaying extractive summaries, content visualization or the combination alongside the interleaved texts. %\footnote{http://resources.trustyou.com/c/wp-present-travel-review-content?x=0MFT5U} %However, while an extractive system selects topic-wise high ranking posts, it may fail to capture the entire argument corresponding to them. Instead, we propose an end-to-end abstractive system which not only avoids error propagation but also hyper-parameters tuning like the number of conversation threads.
\tabref{result_example} shows an output of our hierarchical abstractive system, in which interleaved texts are in the top, and ground-truth and generated summaries in the bottom. \tabref{result_example} also shows the top two post indexes attended by the post-level attention ($\boldsymbol{\gamma}$) while generating those summaries, and they coincide with relevant posts. Similarly, the top 10 indexes (words) of the phrase-level attention ($\boldsymbol{\beta}$) is directly visualized in the table through the color coding matching the generation. The system not only manages to disentangle the interleaved texts but also to generate appropriate abstractive summaries. Meanwhile, $\boldsymbol{\beta}$ provides explainability of the output.

%As interleaving in the table includes abstracts from the same category, e.g. Physical Activities, the interleaving is complex and approximates the real-world conversations. Despite that, the end-to-end hierarchical system tackles the task to a large extent. 
The next step in this research is transfer learning of the hierarchical system trained on the synthetic corpus to real-world examples. %We find models trained for 50k iteration on Pubmed Hard corpora (a=6, b=10, m=2 and n=3) and then fine-tuned on the popular meeting AMI corpus (\cite{7529878bc1a143dbad4fa019e742fdb8}) already reaches competitive results (Rouge-1=39.81, Rouge-2=11.35) against SOA system on AMI (\cite{P18-1062} Rouge-1=37.86 and Rouge-2=07.84). %(e.g. Rouge-1=21.7 and Rouge-2=02.5 of TextRank (\cite{mihalcea2004textrank}), Rouge-1=27.9, Rouge-2=04.10 of FUSION (\cite{mehdad2013abstractive})). 
Further, we aim to modify hier2hier to include some of the recent additions of seq2seq models, e.g., \newcite{P17-1099} pointer mechanism. %, which we leave for future work.

\begin{table}[!t]
\begin{center}
\resizebox{1.0\linewidth}{!}{
%\footnotesize
%\begin{tabular}{p{0.95\linewidth}}%{lp{0.92\linewidth}}
%\begin{tabular}{m{0.018\linewidth}|m{0.982\linewidth}}
\begin{tabular}{m{0.04\linewidth}|m{0.96\linewidth}}
\hline
&\multicolumn{1}{c}{Interleaved Texts}\\
\Tstrut $0$& this study was conducted \boldblue{to evaluate the influence of} excessive \boldblue{sweating} during \boldblue{long-distance} running \boldblue{on} the urinary concentration of \boldblue{caffeine}\ldots\\
$1$& \boldgreen{to assess the effect of} a \boldgreen{program of} supervised \boldgreen{fitness} walking and patient education on functional status , pain , and\ldots\\
%\vdots& \multicolumn{1}{c}{\vdots}\\
$\dots$ & \multicolumn{1}{c}{\dots}\\
$5$& a total of 102 patients with a documented diagnosis of primary osteoarthritis of one or both knees participated\ldots\\
$6$& we \boldred{examined the effects of intensity of training} on ratings of perceived exertion \boldred{(}\ldots\\
%\vdots& \multicolumn{1}{c}{\vdots}\\
$\dots$ & \multicolumn{1}{c}{\dots}\\
\hline
&\multicolumn{1}{c}{GroundTruth/Generation}\\
& caffeine in sport . influence of endurance exercise on the urinary caffeine concentration .\\
0,2\vspace{0.2cm}& \boldblue{effect of excessive [UNK] during [UNK] running on the urinary concentration of caffeine .}\vspace{0.2cm}\\ 
% & \\
& supervised fitness walking in patients with osteoarthritis of the knee . a randomized , controlled trial .\\
1,4\vspace{0.2cm}& \boldgreen{effect of a physical fitness walking on functional status , pain , and pain}\vspace{0.2cm}\\
% & \\
& the effect of training intensity on ratings of perceived exertion .\\
6,8& \boldred{effects of intensity of training on perceived [UNK] in [UNK] athletes .}\\
%&  \multicolumn{1}{c}{\bf6, \bf8}\\
\hline
\end{tabular}
}
\end{center}
\caption{\tablabel{result_example}Interleaved sentences of 3 articles, and corresponding ground-truth and hier2hier generated summaries. The top 2 sentences that were attended ($\boldsymbol{\gamma}$) for the generation are on the left. Additionally, top words ($\boldsymbol{\beta}$) attended for the generation are colored accordingly.}
\end{table}

\section{Conclusion}
We presented an end-to-end trainable hierarchical encoder-decoder architecture with novel hierarchical attention which implicitly disentangles interleaved texts and generates abstractive summaries covering the text threads. The architecture addresses the error propagation and fluency issues that occur in the two-step architectures, and thereby, adding performance gains of 20-40\% on both real-world and synthetic datasets.
\bibliography{topicsum}

\begin{thebibliography}{23}
\expandafter\ifx\csname natexlab\endcsname\relax\def\natexlab#1{#1}\fi

\bibitem[{Aker et~al.(2016)Aker, Paramita, Kurtic, Funk, Barker, Hepple, and
  Gaizauskas}]{aker2016automatic}
Ahmet Aker, Monica Paramita, Emina Kurtic, Adam Funk, Emma Barker, Mark Hepple,
  and Rob Gaizauskas. 2016.
\newblock Automatic label generation for news comment clusters.
\newblock In \emph{Proceedings of the 9th International Natural Language
  Generation Conference}, pages 61--69.

\bibitem[{Anguera et~al.(2012)Anguera, Bozonnet, Evans, Fredouille, Friedland,
  and Vinyals}]{anguera2012speaker}
Xavier Anguera, Simon Bozonnet, Nicholas Evans, Corinne Fredouille, Gerald
  Friedland, and Oriol Vinyals. 2012.
\newblock Speaker diarization: A review of recent research.
\newblock \emph{IEEE Transactions on Audio, Speech, and Language Processing},
  20(2):356--370.

\bibitem[{Bahdanau et~al.(2014)Bahdanau, Cho, and
  Bengio}]{DBLP:journals/corr/BahdanauCB14}
Dzmitry Bahdanau, Kyunghyun Cho, and Yoshua Bengio. 2014.
\newblock \href {http://arxiv.org/abs/1409.0473} {Neural machine translation by
  jointly learning to align and translate}.
\newblock \emph{CoRR}, abs/1409.0473.

\bibitem[{Barker et~al.(2016)Barker, Paramita, Aker, Kurtic, Hepple, and
  Gaizauskas}]{barker2016sensei}
Emma Barker, Monica~Lestari Paramita, Ahmet Aker, Emina Kurtic, Mark Hepple,
  and Robert Gaizauskas. 2016.
\newblock The sensei annotated corpus: Human summaries of reader comment
  conversations in on-line news.
\newblock In \emph{Proceedings of the 17th annual meeting of the special
  interest group on discourse and dialogue}, pages 42--52.

\bibitem[{Cheng and Lapata(2016)}]{cheng2016neural}
Jianpeng Cheng and Mirella Lapata. 2016.
\newblock Neural summarization by extracting sentences and words.
\newblock \emph{arXiv preprint arXiv:1603.07252}.

\bibitem[{Chopra et~al.(2016)Chopra, Auli, and
  Rush}]{DBLP:conf/naacl/ChopraAR16}
Sumit Chopra, Michael Auli, and Alexander~M. Rush. 2016.
\newblock \href {http://aclweb.org/anthology/N/N16/N16-1012.pdf} {Abstractive
  sentence summarization with attentive recurrent neural networks}.
\newblock In \emph{{NAACL} {HLT} 2016, The 2016 Conference of the North
  American Chapter of the Association for Computational Linguistics: Human
  Language Technologies, San Diego California, USA, June 12-17, 2016}, pages
  93--98. The Association for Computational Linguistics.

\bibitem[{Hsu et~al.(2018)Hsu, Lin, Lee, Min, Tang, and Sun}]{P18-1013}
Wan-Ting Hsu, Chieh-Kai Lin, Ming-Ying Lee, Kerui Min, Jing Tang, and Min Sun.
  2018.
\newblock \href {http://aclweb.org/anthology/P18-1013} {A unified model for
  extractive and abstractive summarization using inconsistency loss}.
\newblock In \emph{Proceedings of the 56th Annual Meeting of the Association
  for Computational Linguistics (Volume 1: Long Papers)}, pages 132--141.
  Association for Computational Linguistics.

\bibitem[{Jiang et~al.(2018)Jiang, Chen, Chen, and Wang}]{N18-1164}
Jyun-Yu Jiang, Francine Chen, Yan-Ying Chen, and Wei Wang. 2018.
\newblock \href {https://doi.org/10.18653/v1/N18-1164} {Learning to disentangle
  interleaved conversational threads with a siamese hierarchical network and
  similarity ranking}.
\newblock In \emph{Proceedings of the 2018 Conference of the North American
  Chapter of the Association for Computational Linguistics: Human Language
  Technologies, Volume 1 (Long Papers)}, pages 1812--1822. Association for
  Computational Linguistics.

\bibitem[{Jing et~al.(2018)Jing, Xie, and Xing}]{P18-1240}
Baoyu Jing, Pengtao Xie, and Eric Xing. 2018.
\newblock \href {http://aclweb.org/anthology/P18-1240} {On the automatic
  generation of medical imaging reports}.
\newblock In \emph{Proceedings of the 56th Annual Meeting of the Association
  for Computational Linguistics (Volume 1: Long Papers)}, pages 2577--2586.
  Association for Computational Linguistics.

\bibitem[{Krause et~al.(2017)Krause, Johnson, Krishna, and
  Fei-Fei}]{krause2016paragraphs}
Jonathan Krause, Justin Johnson, Ranjay Krishna, and Li~Fei-Fei. 2017.
\newblock A hierarchical approach for generating descriptive image paragraphs.
\newblock In \emph{Computer Vision and Patterm Recognition (CVPR)}.

\bibitem[{Li et~al.(2015)Li, Luong, and Jurafsky}]{P15-1107}
Jiwei Li, Thang Luong, and Dan Jurafsky. 2015.
\newblock \href {https://doi.org/10.3115/v1/P15-1107} {A hierarchical neural
  autoencoder for paragraphs and documents}.
\newblock In \emph{Proceedings of the 53rd Annual Meeting of the Association
  for Computational Linguistics and the 7th International Joint Conference on
  Natural Language Processing (Volume 1: Long Papers)}, pages 1106--1115.
  Association for Computational Linguistics.

\bibitem[{Ma et~al.(2012)Ma, Sun, Yuan, and Cong}]{ma2012topic}
Zongyang Ma, Aixin Sun, Quan Yuan, and Gao Cong. 2012.
\newblock Topic-driven reader comments summarization.
\newblock In \emph{Proceedings of the 21st ACM international conference on
  Information and knowledge management}, pages 265--274. ACM.

\bibitem[{McCowan et~al.(2005)McCowan, Carletta, Kraaij, Ashby, Bourban, Flynn,
  Guillemot, Hain, Kadlec, Karaiskos, Kronenthal, Lathoud, Lincoln, Lisowska,
  Post, Reidsma, and Wellner}]{7529878bc1a143dbad4fa019e742fdb8}
I.~McCowan, J.~Carletta, W.~Kraaij, S.~Ashby, S.~Bourban, M.~Flynn,
  M.~Guillemot, T.~Hain, J.~Kadlec, V.~Karaiskos, M.~Kronenthal, G.~Lathoud,
  M.~Lincoln, A.~Lisowska, W.~Post, Dennis Reidsma, and P.~Wellner. 2005.
\newblock The ami meeting corpus.
\newblock In \emph{Proceedings of Measuring Behavior 2005, 5th International
  Conference on Methods and Techniques in Behavioral Research}, pages 137--140.
  Noldus Information Technology.

\bibitem[{Nallapati et~al.(2017)Nallapati, Zhai, and Zhou}]{AAAI1714636}
Ramesh Nallapati, Feifei Zhai, and Bowen Zhou. 2017.
\newblock \href
  {https://aaai.org/ocs/index.php/AAAI/AAAI17/paper/view/14636/14080}
  {Summarunner: A recurrent neural network based sequence model for extractive
  summarization of documents}.
\newblock In \emph{AAAI Conference on Artificial Intelligence}.

\bibitem[{Nallapati et~al.(2016)Nallapati, Zhou, dos Santos,
  G{\"{u}}l{\c{c}}ehre, and Xiang}]{DBLP:conf/conll/NallapatiZSGX16}
Ramesh Nallapati, Bowen Zhou, C{\'{\i}}cero~Nogueira dos Santos, {\c{C}}aglar
  G{\"{u}}l{\c{c}}ehre, and Bing Xiang. 2016.
\newblock \href {http://aclweb.org/anthology/K/K16/K16-1028.pdf} {Abstractive
  text summarization using sequence-to-sequence rnns and beyond}.
\newblock In \emph{Proceedings of the 20th {SIGNLL} Conference on Computational
  Natural Language Learning, CoNLL 2016, Berlin, Germany, August 11-12, 2016},
  pages 280--290. {ACL}.

\bibitem[{Noh et~al.(2017)Noh, Araujo, Sim, Weyand, and Han}]{noh2017large}
Hyeonwoo Noh, Andre Araujo, Jack Sim, Tobias Weyand, and Bohyung Han. 2017.
\newblock Large-scale image retrieval with attentive deep local features.
\newblock In \emph{Proceedings of the IEEE International Conference on Computer
  Vision}, pages 3456--3465.

\bibitem[{Rush et~al.(2015)Rush, Chopra, and Weston}]{DBLP:conf/emnlp/RushCW15}
Alexander~M. Rush, Sumit Chopra, and Jason Weston. 2015.
\newblock \href {http://aclweb.org/anthology/D/D15/D15-1044.pdf} {A neural
  attention model for abstractive sentence summarization}.
\newblock In \emph{Proceedings of the 2015 Conference on Empirical Methods in
  Natural Language Processing, {EMNLP} 2015, Lisbon, Portugal, September 17-21,
  2015}, pages 379--389. The Association for Computational Linguistics.

\bibitem[{See et~al.(2017)See, Liu, and Manning}]{P17-1099}
Abigail See, Peter~J. Liu, and Christopher~D. Manning. 2017.
\newblock \href {https://doi.org/10.18653/v1/P17-1099} {Get to the point:
  Summarization with pointer-generator networks}.
\newblock In \emph{Proceedings of the 55th Annual Meeting of the Association
  for Computational Linguistics (Volume 1: Long Papers)}, pages 1073--1083.
  Association for Computational Linguistics.

\bibitem[{Shang et~al.(2018)Shang, Ding, Zhang, Tixier, Meladianos,
  Vazirgiannis, and Lorr{\'e}}]{P18-1062}
Guokan Shang, Wensi Ding, Zekun Zhang, Antoine Tixier, Polykarpos Meladianos,
  Michalis Vazirgiannis, and Jean-Pierre Lorr{\'e}. 2018.
\newblock \href {http://aclweb.org/anthology/P18-1062} {Unsupervised
  abstractive meeting summarization with multi-sentence compression and
  budgeted submodular maximization}.
\newblock In \emph{Proceedings of the 56th Annual Meeting of the Association
  for Computational Linguistics (Volume 1: Long Papers)}, pages 664--674.
  Association for Computational Linguistics.

\bibitem[{Tan et~al.(2017)Tan, Wan, and Xiao}]{tan2017neural}
Jiwei Tan, Xiaojun Wan, and Jianguo Xiao. 2017.
\newblock From neural sentence summarization to headline generation: A
  coarse-to-fine approach.
\newblock In \emph{IJCAI}, pages 4109--4115.

\bibitem[{Teichmann et~al.(2019)Teichmann, Araujo, Zhu, and
  Sim}]{teichmann2019detect}
Marvin Teichmann, Andre Araujo, Menglong Zhu, and Jack Sim. 2019.
\newblock Detect-to-retrieve: Efficient regional aggregation for image search.
\newblock In \emph{Proceedings of the IEEE Conference on Computer Vision and
  Pattern Recognition}, pages 5109--5118.

\bibitem[{Verberne et~al.(2018)Verberne, Krahmer, Hendrickx, Wubben, and van
  Den~Bosch}]{verberne2018creating}
Suzan Verberne, Emiel Krahmer, Iris Hendrickx, Sander Wubben, and Antal van
  Den~Bosch. 2018.
\newblock Creating a reference data set for the summarization of discussion
  forum threads.
\newblock \emph{Language Resources and Evaluation}, pages 1--23.

\bibitem[{Wang and Oard(2009)}]{wang2009context}
Lidan Wang and Douglas~W Oard. 2009.
\newblock Context-based message expansion for disentanglement of interleaved
  text conversations.
\newblock In \emph{Proceedings of human language technologies: The 2009 annual
  conference of the North American chapter of the association for computational
  linguistics}, pages 200--208. Association for Computational Linguistics.

\end{thebibliography}
\bibliographystyle{acl_natbib}

\newpage

\appendix
\section*{Appendix}

\section{Interleave Algorithm}
In \algref{interleave_algo}, $\textsc{Interleave}$ takes a set of concatenated abstracts and titles, $\textit{C} = \langle\textit{A}_1;\textit{T}_1,\ldots,\textit{A}_{|C|};\textit{T}_{|C|}\rangle$, minimum, $a$, and maximum, $b$, number of abstracts to interleave, and minimum, $m$, and maximum, $n$, number of sentences in a source, and then returns a set of concatenated interleaved texts and summaries. $\textsc{window}$ takes a sequence of texts, $\textit{X}$, and returns a window iterator of size $\frac{|\mathit{X}|-\textit{w}}{t}+1$, where $\textit{w}$ and $\textit{t}$ are window size and sliding step respectively. $\textit{window}$ reuses elements of $\textit{X}$, and therefore, enlarges the corpus size. Notations $\mathcal{U}$ refers to a uniform sampling, $\left[\cdot\right]$ to array indexing, and $\textsc{Reverse}$ to reversing an array.

\label{app:algo}
\begin{algorithm}[!h]
\footnotesize
\caption{Interleaving Algorithm}\label{alg:interleave_algo}
\begin{algorithmic}[1]
\Procedure{Interleave}{$\textit{C}, \textit{a}, \textit{b}, \textit{m}, \textit{n}$}
\State $\hat{\textit{C}},\textit{Z}\gets \textsc{window}(\textit{C}, w=b, t=1)$, Array()
\While {$\hat{\textit{C}} \neq \emptyset$}
	\State $\textit{C}^\prime, \textit{A}^\prime, \textit{T}^\prime, \textit{S} \gets \hat{\textit{C}}.\textsc{Next}()$, Array(), Array(), $\{\}$
	\State ${\textit{r}}$ $\sim$ $\mathcal{U}$($\textit{a}, \textit{b}$)	
	\For {$\textit{j}$ = 1 to $\textit{r}$} \Comment{Selection}
		\State $\textit{A}, \textit{T}  \gets \hat{{\textit{C}}}[\textit{j}]$
		\State $\textit{T}^\prime.\textsc{Add}(\textit{T}$)
		\State ${\textit{q}}$ $\sim$ $\mathcal{U}$($\textit{m}, \textit{n}$)			
		\State $\textit{A}^\prime.\textsc{Add}(\textit{A}$[1:$\textit{q}$])
		\State $\textit{S} \gets \textit{S}\cup\lbrace j_{\times q}\rbrace$
\EndFor
\State $\hat{\textit{A}}, \hat{\textit{T}} \gets$ Array(), Array()	
	\For {1 to $|{\textit{S}}|$}	\Comment{Interleaving}
		\State $\textit{k} \gets \mathcal{U}(\textit{S})$
		\State $\textit{S} \gets \textit{S}\backslash{\textit{k}}$		
		\State $\textit{I} \gets \textsc{Reverse}(\textit{A}^\prime$[$\textit{k}]).\textsc{pop}()$
		\State $\hat{\textit{A}}.\textsc{Add}(\textit{I}$)		
		\State $\textit{J} \gets \textit{T}^{\prime}$[$\textit{k}$]
		\If {$\textit{J} \not\in \hat{\textit{T}}$}:		
			\State $\hat{\textit{T}}.\textsc{Add}(\textit{J}$)
		\EndIf					
	\EndFor
	\State $\textit{Z}.\textsc{Add}(\hat{\textit{A}}$;$\hat{\textit{T}}$)
\EndWhile
\State \Return $\textit{Z}$
\EndProcedure
\end{algorithmic}
\end{algorithm}

\section{Parameters}
For the word-to-word encoder, the steps are limited to 20, while the steps in the word-to-word decoder are limited to 15. The steps in the post-to-post encoder and thread-to-thread decoder depend on the corpus type, e.g., Medium has 15 steps in post-to-post and 3 steps in thread-to-thread. In seq2seq experiments, the source is flattened, and therefore, the number of steps in the source is limited to 300. We initialized all weights, including word embeddings, with a random normal distribution with mean 0 and standard deviation 0.1. The embedding vectors and hidden states of the encoder and decoder in the models are set to dimension 100. Texts are lowercased. The vocabulary size is limited to 8000 and 15000 for Pubmed and Stack Exchange corpora respectively. We pad short sequences with a special token, $\langle PAD\rangle$. We use Adam (Kingma et al. 2014)%\cite{DBLP:journals/corr/KingmaB14} 
with an initial learning rate of .0001 and batch size of 64 for training. % and numbers are replaced by the special symbol $\%$. 

\section{Training Loss}

\begin{figure}[h!]
\centering
\resizebox{0.65\linewidth}{!}{
\includegraphics[width=0.99\textwidth]{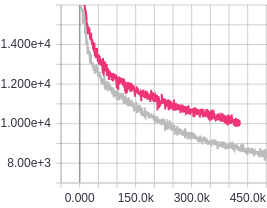}
}
\caption{Running average training loss between seq2seq (pink) and hier2hier (gray) for Stack Exchange Hard corpus.}
\figlabel{hier_perf_comp2}
\end{figure}

\section{Examples}
\begin{table*}[htbp!]
\begin{center}
\resizebox{0.99\linewidth}{!}{
\footnotesize
%\begin{tabular}{p{0.95\linewidth}}%{lp{0.92\linewidth}}
\begin{tabular}{p{0.54\linewidth}|p{0.45\linewidth}}
\hline
\ding{51} botulinum toxin a is effective for treatment\ldots&\multirow{4}{5.5cm}{\ding{51} prospective randomised controlled trial comparing trigone-sparing versus trigone-including intradetrusor injection of abobotulinumtoxina for refractory idiopathic detrusor overactivity.}\\
\ding{51} the trigone is generally spared because of the theoretical\ldots&\\
\ding{51} evaluate efficacy and safety of trigone-including .\ldots&\\
\ding{70} most methadone-maintained injection drug users \ldots&\\
%\vdots & \multicolumn{1}{c}{\vdots}\\
%\multicolumn{1}{c|}{\ldots}&\\
\ding{81} gender-related differences in the incidence of bleeding\ldots&\\
\ding{81} we studied patients with stemi receiving fibrinolysis\ldots&\\
\ding{70} physicians may be reluctant to treat hcv in idus because \ldots&\multirow{3}{5.8cm}{\ding{70} rationale and design of a randomized controlled trial of directly observed hepatitis c treatment delivered in methadone clinics.}\\
\ding{81} outcomes included moderate or severe bleeding defined 
\ldots&\\
\ding{70} optimal hcv management approaches for idus remain
\ldots&\\
\ding{81} moderate or severe bleeding was 1.9-fold higher 
\ldots&\multirow{2}{5cm}{\ding{81} comparison of incidence of bleeding and mortality of men versus women with st-elevation myocardial infarction treated with fibrinolysis.}\\
\ding{70} we are conducting a randomized controlled trial in a network\ldots&\\
\ding{81} bleeding remained higher in women even after adjustment 
\ldots&\\
\hline
\end{tabular}}
\end{center}
\caption{\tablabel{example_dataset_a1}The left rows contain interleaving of 3 articles with 2 to 5 sentences and the right rows contain their interleaved titles. Associated sentences and titles are depicted by similar symbols.}
\end{table*}

\begin{table*}[htbp!]
\begin{center}
\resizebox{0.99\linewidth}{!}{
\footnotesize
%\begin{tabular}{p{0.95\linewidth}}%{lp{0.92\linewidth}}
\begin{tabular}{p{0.55\linewidth}|p{0.45\linewidth}}
\hline
\ding{51} the effects of short-course antiretrovirals given to\ldots&\multirow{4}{5.5cm}{\ding{51} hiv-1 persists in breast milk cells despite antiretroviral treatment to prevent mother-to-child transmission.}\\
\ding{81} good adherence is essential for successful antiretroviral\ldots&\\
\ding{51} women in kenya received short-course zidovudine ( zdv )\ldots&\\
\ding{51} breast milk samples were collected two to three times weekly.\ldots&\\
\ding{70} the present primary analysis of antiretroviral therapy with\ldots&\multirow{3}{5.8cm}{\ding{81} patterns of individual and population-level adherence to antiretroviral therapy and risk factors for poor adherence in the first year of the dart trial in uganda and zimbabwe.}\\
\ding{81} this was an observational analysis of an open multicenter\ldots&\\
\ding{70} patients with hiv-1 rna at least 5000 copies/ml were\ldots&\\
\ding{81} at 4-weekly clinic visits , art drugs were provided and \ldots&\\
\ding{70} the primary objective was to demonstrate non-inferiority\ldots&\\
\ding{81} viral load response was assessed in a subset of patients\ldots&\multirow{2}{5cm}{\ding{70} efficacy and safety of once-daily darunavir/ritonavir versus lopinavir/ritonavir in treatment-naive hiv-1-infected patients at week 48.}\\
\ding{168} we explored the link between serum alpha-fetoprotein levels\ldots&\\
\ding{81} drug possession ratio ( percentage of drugs taken between\ldots&\\
\ding{168} a low alpha-fetoprotein level ( $<$ 5.0 ng/ml ) was an\ldots&\\
\ding{70} six hundred and eighty-nine patients were randomized\ldots&\\
\ding{70} at 48 weeks , 84 \% of drv/r and 78 \% of lpv/r patients\ldots&\multirow{2}{5cm}{\ding{168} serum alpha-fetoprotein predicts virologic response to hepatitis c treatment in hiv coinfected patients.}\\
\ding{51} hiv-1 dna was quantified by real-time pcr .\ldots&\\
\ding{168} serum alpha-fetoprote in measurement should be integrated \ldots&\\
\hline
\end{tabular}
}
\end{center}
\caption{\tablabel{example_dataset_a2}The left rows contain interleaving of 4 articles with 2 to 5 sentences and the right rows contain their interleaved titles. Associated sentences and titles are depicted by similar symbols.}
\end{table*}

\begin{table}[!h]
\begin{center}
\resizebox{0.95\linewidth}{!}{
\begin{tabular}{m{0.02\linewidth}|m{0.95\linewidth}}
\hline
&\multicolumn{1}{c}{Interleaved Texts}\\
\Tstrut $0$& \boldblue{botulinum} \boldblue{toxin} a is \boldblue{effective} for \boldblue{treatment} of \boldblue{idiopathic} \boldblue{detrusor} overactivity ( [UNK] )\\
$1$& the [UNK] is generally [UNK] because of the theoretical risk of [UNK] reflux ( [UNK] ) , although studies assessing\ldots\\
$\dots$ & \multicolumn{1}{c}{\dots}\\
$3$& \boldgreen{most} \boldgreen{[UNK]} \boldgreen{injection} \boldgreen{drug} \boldgreen{users} ( \boldgreen{idus} ) have been infected with hepatitis c virus ( hcv ) , but\ldots\\
$4$&\boldred{[UNK]} \boldred{differences} in the \boldred{incidence} of \boldred{bleeding} and its relation to subsequent mortality in patients with st-segment elevation myocardial infarction\ldots\\
%\vdots& \multicolumn{1}{c}{\vdots}\\
$\dots$ & \multicolumn{1}{c}{\dots}\\
$8$& \boldred{optimal} hcv management approaches for idus remain unknown .\ldots\\
%\vdots& \multicolumn{1}{c}{\vdots}\\
$\dots$ & \multicolumn{1}{c}{\dots}\\
\hline
&\multicolumn{1}{c}{GroundTruth/Generation}\\
& prospective randomised controlled trial comparing trigone-sparing versus trigone-including intradetrusor injection of abobotulinumtoxina for refractory idiopathic detrusor overactivity.\\
0,1& \boldblue{efficacy of [UNK] [UNK] in patients with idiopathic detrusor overactivity : rationale , design}\\ 
 & \\
%&  \multicolumn{1}{c}{\bf0, \bf2}\\
%\Tstrut \multirow{2}*{$1,4$}
& rationale and design of a randomized controlled trial of directly observed hepatitis c treatment delivered in methadone clinics.\\
3,4& \boldgreen{validation of a point-of-care hepatitis injection drug injection drug , hcv medication , and}\\
 & \\
%&  \multicolumn{1}{c}{\bf1, \bf4}\\
%\Tstrut \multirow{2}*{$6,8$}
& comparison of incidence of bleeding and mortality of men versus women with st-elevation myocardial infarction treated with fibrinolysis .\\
4,8& \boldred{subgroup analysis of patients with st-elevation myocardial infarction with st-elevation myocardial infarction .}\\
%&  \multicolumn{1}{c}{\bf6, \bf8}\\
\hline
\end{tabular}
}
\end{center}
\caption{\tablabel{result_example_a1}Interleaved sentences of 3 articles, and corresponding ground-truth and hier2hier generated summaries. The top 2 sentences that were attended ($\boldsymbol{\gamma}$) for the generation are on the left. Additionally, top words ($\boldsymbol{\beta}$) attended for the generation are colored accordingly.}
\end{table}

\begin{table}[!h]
\begin{center}
\resizebox{0.95\linewidth}{!}{
\begin{tabular}{m{0.04\linewidth}|m{0.95\linewidth}}
\hline
&\multicolumn{1}{c}{Interleaved Texts}\\
\Tstrut $0$& the \boldblue{effects} of \boldblue{short-course} \boldblue{antiretrovirals} \boldblue{given} to \boldblue{reduce} \boldblue{mother-to-child} \boldblue{transmission} ( [UNK] ) on temporal patterns of [UNK] hiv-1 rna\\
$1$& \boldgreen{good} \boldgreen{adherence} \boldgreen{is} \boldgreen{essential} \boldgreen{for} \boldgreen{successful} \boldgreen{antiretroviral} \boldgreen{therapy} ( \boldgreen{art} ) provision , but simple measures have rarely been validated\ldots\\
$2$& women in kenya received short-course zidovudine ( zdv ) , single-dose nevirapine ( [UNK] ) , combination [UNK] or short-course\ldots\\
$3$& breast milk samples were collected two to three times weekly for 4-6 weeks .\ldots\\
$4$&the \boldred{present} \boldred{primary} \boldred{analysis} of \boldred{antiretroviral} \boldred{therapy} \boldred{with} \boldred{
[UNK]} \boldred{examined} \boldred{in} naive subjects ( [UNK] ) compares the efficacy and\ldots\\
%\vdots& \multicolumn{1}{c}{\vdots}\\
$\dots$ & \multicolumn{1}{c}{\dots}\\
$10$& we \boldyellow{explored} \boldyellow{the} \boldyellow{link} \boldyellow{between} \boldyellow{serum} \boldyellow{[UNK]} levels and virologic response in [UNK] [UNK] c virus coinfected patients .\ldots\\
%\vdots& \multicolumn{1}{c}{\vdots}\\
$\dots$ & \multicolumn{1}{c}{\dots}\\
\hline
&\multicolumn{1}{c}{GroundTruth/Generation}\\
& hiv-1 persists in breast milk cells despite antiretroviral treatment to prevent mother-to-child transmission .\\
0,2& \boldblue{impact of hiv-1 persists on hiv-1 rna in human immunodeficiency virus-infected individuals with hiv-1}\\ 
 & \\
& patterns of individual and population-level adherence to antiretroviral therapy and risk factors for poor adherence in the first year of the dart trial in uganda and zimbabwe .\\
1,3& \boldgreen{impact of a antiretroviral treatment algorithm on adherence to antiretroviral therapy in [UNK] ,}\\
 & \\
& efficacy and safety of once-daily darunavir/ritonavir versus lopinavir/ritonavir in treatment-naive hiv-1-infected patients at week 48 .\\
4,2& \boldred{a randomized trial of [UNK] versus [UNK] in treatment-naive hiv-1-infected patients with hiv-1 infection}\\
& \\
& serum alpha-fetoprotein predicts virologic response to hepatitis c treatment in hiv coinfected patients .\\
\multirow{2}{1cm}{10,12}& \boldyellow{predicting virologic response in [UNK] coinfected patients coinfected with hiv-1 : a [UNK] randomized}\\
\hline
\end{tabular}
}
\end{center}
\caption{\tablabel{result_example_a2}Interleaved sentences of 4 articles, and corresponding ground-truth and hier2hier generated summaries. The top 2 sentences that were attended ($\boldsymbol{\gamma}$) for the generation are on the left. Additionally, top words ($\boldsymbol{\beta}$) attended for the generation are colored accordingly.}
\end{table}

\end{document}